\pdfoutput=1
\documentclass{article}

\usepackage{arxiv}

\usepackage[utf8]{inputenc} 
\usepackage[T1]{fontenc}    
\usepackage{hyperref}       
\usepackage{url}            
\usepackage{booktabs}       
\usepackage{amsfonts}       
\usepackage{nicefrac}       
\usepackage{microtype}      
\usepackage{lipsum}
\usepackage{graphicx}
\graphicspath{ {./images/} }
\usepackage{graphicx}
\graphicspath{ {./images/} }
\usepackage{doi}
\usepackage{amsmath}
\usepackage{algorithm}
\usepackage{algpseudocode}
\usepackage{mwe}
\usepackage{subfig}
\usepackage{array}
\usepackage{pdfpages}
\usepackage[numbers]{natbib}

\title{Rectified Gaussian kernel multi-view k-means clustering}

\author{\href{https://orcid.org/0009-0000-6184-829X}{\includegraphics[scale=0.06]{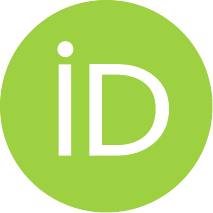}\hspace{1mm}Kristina P. ~Sinaga} \\
  Independent Researcher\\
  \texttt{kristinasinaga41@gmail.com} \\
}

\begin{document}
\maketitle
\begin{abstract}
In this paper, we show two new variants of multi-view k-means (MVKM) algorithms to address multi-view data. The general idea is to outline the distance between \textit{h}-th view data points $x_i^h$ and \textit{h}-th view cluster centers $a_k^h$ in a different manner of centroid-based approach. Unlike other methods, our proposed methods learn the multi-view data by calculating the similarity using Euclidean norm in the space of Gaussian-kernel, namely as multi-view k-means with exponent distance (MVKM-ED). By simultaneously aligning the stabilizer parameter $p$ and kernel coefficients $\beta^h$, the compression of Gaussian-kernel based weighted distance in Euclidean norm reduce the sensitivity of MVKM-ED. To this end, this paper designated as Gaussian-kernel multi-view k-means (GKMVKM) clustering algorithm. Numerical evaluation of five real-world multi-view data demonstrates the robustness and efficiency of our proposed MVKM-ED and GKMVKM approaches.
\keywords{Multi-view clustering \and k-means  \and Exponential distance  \and Gaussian-kernel.}
\end{abstract}


\section{Introduction}
The goal of data analysis is basically to investigate unusual behaviors and make an improvement or accurate strategy at reliable costs on the next action. In general, data analysis can be processed in supervised, semi-supervised, and unsupervised manner. One of the most well-known unsupervised manners is clustering technique. Common clustering techniques includes k-means  \cite{macqeen1967some}, fuzzy c-means (FCM) \cite{bezdek1984fcm}, and possibilistic c-means (PCM) \cite{krishnapuram1993possibilistic} algorithms. These clustering techniques had been widely extended and implemented in many real-world applications such as solving industrial problem \cite{cheng2014using, kolouri2017optimal, mourelo2016optimization}, medical diagnosis \cite{ng2006medical, khanmohammadi2017improved}, image segmentation \cite{lupacscu2011automatic, dhanachandra2015image, altini2021segmentation}, smart city or IoT data problem \cite{liu2020privacy}, etc. One of the effective approaches with huge benefits and most robust linearly promoted to the rapid development of technology is well-known as to be multi-view learning (MVL). 

\paragraph{} Unlike single-view learning (SVL), MVL utilize multiple resources with diversity information of the same observation. MVL techniques focuses on representing or decomposing MV data into a low-rank space before clustering processes are usually based on Non-Negative Matrix Factorization (NMF) \cite{luong2022multi} and tensor-based approaches. Both learns the MV data by revealing their local geometric structure and transforming them into a new lower space. These NMF-based and tensor-based approaches are working well especially on highly sparse and higher-order dataset. However, MVL-NMF-based approach somehow produces undesirable or poor clustering results due to the ability of sigmoid or manifold learning to produce a non-singularity matrix is limited. On the other hand, MVL-Tensor-based approach required highly expensive costs but worth it to solve a higher-order problem. 

\paragraph{} MVL with non-orthogonal-based approach search the optimal clustering results by processing all data matrices directly into account. The retrieval MV data with non-orthogonal-based approaches including cluster centers-based approach, collaborative-based approach, etc. Most of non-orthogonal-based approaches used the Euclidean function to estimate the similarities between data points across views. We notice that non-orthogonality-MVL-based technique is one of the best approaches to solve MV problem. However, we realized that there is a less research effort has been made to consider an alternative way for discriminating process in clustering techniques. 

\paragraph{} In this paper, our proposed approach is to generalize the cluster-based Euclidean distance into a new learning metric to better cover the MV problem from the view point of k-means clustering algorithm. In such cases, we present new Euclidean norm-based-distances to ensure the representations of similar samples to be close and dissimilar samples to be distant. We proposed four objective functions starting from the very basic one of considering Euclidean norm distance, and its extension into new transform learning Gaussian-Kernel-based distance with (out) kernel coefficients $\beta^h$ and $p$ stabilizer parameter.

\section{Related Works}
\paragraph{Multi-View Learning (MVL). }MVL is one of the most robust approaches to solve the heterogeneous of multiple resources data. It had been received and attracted the attention of researcher from different fields. The growth of massive data coming from multiple devices making it possible that the information of the same observation may occur bias and including the uninformative feature during clustering processes. Hence, the structuring process are varied, depends on the objective function and its constraints. In this section, we will briefly review the distance-based clustering approaches in single-view and multi-view k-means clustering algorithms. 

\paragraph{k-means.}
k-means is a popular general-purpose and the simplest procedure clustering but powerful to reveal a pattern of one input data. k-means was introduced by MacQueen in 1967 based on the idea of Lloyd’ algorithm. The original k-means is designed works well to address a data with spherical shapes by taking the weighted minimum distance between data points and its centers. Let $i$ be the indices of data points $x$, $c$ is defined as the number of clusters, and the cluster centers of $j$-th feature component in $k$-th cluster is denoted as  $a_{kj}$. Thus, the objective function of k-means can be expressed as below.

\begin{equation}
J_{\text{k-means}} (U,A) = \sum_{i=1}^{n} \sum_{k=1}^c \mu_{ik} \|x_i - a_k\|^2
\label{eqn:k-means}
\end{equation}

The original k-means clustering is sensitive to initialization and the performances often distracted by noises. In such cases, Sinaga and Yang \cite{sinaga2020unsupervised} introduced a new concept of free initialization to k-means by adding a new term of mixing proportions in the objective function, called as unsupervised k-means clustering (U-k-means). 

\paragraph{Alternative k-means Clustering.} Wu and Yang naming their contribution in k-means clustering as an alternative hard c-means (AHCM) by introducing a new metric to replace the Euclidean norm in k-means clustering with an exponent weighted distance function in Euclidean space \cite{wu2002alternative}. Their new procedures are efficient to handle a noisy environment with promising accuracies. Recognizing the potential of AHCM, Chang-Chien et al. \cite{chang2021gaussian} elaborate the exponent distance function in Wu and Yang \cite{wu2002alternative} to generalize the HCM or K-Means with Gaussian-kernel c-means clustering, called as Gaussian-kernel hard c-means (GK-HCM). These AHCM and GK-HCM are failed to address multiple features data as its procedures are designed to handle a single representation data.

\paragraph{Learning-Distance-Based Approaches.}  In this paper, the potential of new alternative k-means clustering algorithm to Gaussian space-based weighted distance as new efficient and promising ways to analyze multi-view data by quantifying its similarities are inspired by the works of Yang and Wu \cite{wu2002alternative}, Maaten et al. \cite{van2008visualizing} and Fukunaga and Kasai \cite{fukunaga2021wasserstein}. Yang and Wu introduced an alternative technique into an optimization problem by using a self-organize similarity-based clustering with $\beta$ as a normalized parameter and $\gamma$ as a power parameter to control the effect of  $\beta$. Given data point $x_i$ and cluster centers $a_k$, then the self-organize in the proposed similarity-based clustering (SCM) proposed by Yang and Wu \cite{fukunaga2021wasserstein} can be expressed in the following way.

\begin{equation}
S(x_i, a_k) = \text{exp} \bigg( - \frac{\|x_i  - a_k\|^2} {\beta}\bigg)
\end{equation}

Like SCM, the \textit{t} stochastic neighbor embedding (t-SNE) proposed by Maaten et al. \cite{van2008visualizing} used the same similarities concept as Yang and Wu\cite{wu2002alternative}. They enable the mapping of a high dimensionality data into a two or three-dimensional, such as converting the high-dimensional distance between data points into conditional probabilities that represent similarities. Given two data points $x_i$ and $x_j$ , then the conditional probabilities to measure the similarities between those two data points in t-SNE can be expressed in the following way.

\begin{equation}
P(x_j | x_i) =  \bigg(  \frac{\text{exp} (-\|x_i  - x_j\|^2 / 2 \alpha_i^2)} {\sum_{i=1}^n \text{exp} (-\|x_i  -  x_j\|^2 / 2 \alpha_i^2)}\bigg)
\end{equation}

where $\alpha_i$ is the variance of the Gaussian that is centered on data point $x_i$. On the other hand, Wasserstein distance (WD) or also known as Kantorovich-Rubinstein was firstly introduced by Vaserstein in 1969 \cite{vaserstein1969markov}. The WD appeared as a probabilistic-based approach for solving the Markov processes on large system of automata. In the literature, Mémoli \cite{memoli2011gromov} lined up the Wasserstein distances for matching object purposes. Works on distance-based approach to improve the robustness and accuracy in Gaussian distribution with Wasserstein metric is done by Salmona et al. \cite{delon2022gromov}. Imaizumi et al. \cite{imaizumi2022hypothesis} investigates the ability of Wasserstein Distance to analyze general dimension in a data. To integrate the similarity-based clustering (SCM) with conditional probabilities in multi-view k-means clustering, we are adopting similarities approach as new unique and adaptive ways to reveal patterns of one multi-view data. 

\paragraph{Multi-view K-Means Clustering.} The goal of multi-view clustering (MVC) is to group certain number of data instances of multiple data features representations simultaneously into a specified number of clusters. A general distance-based clustering algorithm like Euclidean distance is the most popular function to map a group of data points into different clusters \cite{macqeen1967some, likas2003global}. An automated two-level variable weighting such as two-level variable weighting clustering algorithm (TW-k-means) proposed by Chen et al. \cite{chen2011tw} is an extension of weighted k-means algorithm \cite{huang2005automated} which defined the distribution of data points within one cluster by Euclidean-distance-correlation based function. Similarly, a weighted multi-view clustering with feature selection (WMCFS) proposed by Xu et al. \cite{xu2016weighted}, a simultaneous weighting on views and features (SWVF) by Jiang et al. \cite{jiang2016multi}, and a two-level weighted collaborative k-means by Zhang et al. \cite{zhang2018tw} are used an Euclidean-distance-correlation based function. These multi-view k-means clustering methods measured the distance between the data points and its cluster centers using Euclidean norm. To unlock a new potential of k-means to benefit processing multiple features as input data, we are investigating a new possibility to boost the performance of clustering by elaborating Gaussian-kernel-distance-based procedure into multi-view k-means clustering. To learn next more efficiently, we design a Gaussian-kernel-based distance under a $p$ stabilizer parameter and kernel coefficients $\beta^h$.

\section{The Proposed Methods}
\subsection{Notation}
Let us define a dataset $X \in \mathbb{R}^{n \times D}$ with $X^h = \{ x_1^{d_1} ,\ldots, x_n^{d_h} \}$ and $\sum_{h=1}^s d_h = D$. $x_i^{d_h}$ represents the $i$th points of $x$ in $h$th view. A partition $\mu$ in $X$ is a set of disjoint clusters that partitions $X$ into $c$ clusters across views $h$ such as $\mu = [\mu_{ik}]_{n \times c}$, $\mu_{ik} \in \{0, 1\}$ and $\sum_{k=1}^c \mu_{ik}^h = 1$. $v$ is a view weight vector with  $v_h \in [0,1]$ and $\sum_{h=1}^s v_h = 1$. We use cluster centers $k$ of $h$th view as $A^h$, where $A^h = [a_{kj}^h]_{c \times d_h}$.

\subsection{The Multi-view k-means}

The objective function of k-means in Eq. \ref{eqn:k-means} is disabled to address a data with multiple features representation. To enable the representation of multiple features data during the clustering processes, we need to consider additional variable view $v$ such as different view may contains different information.  A different feature-view representation of the same observation within one data can be designed in two ways such as treat all data views equally importance and non-equally importance. In this paper, we assume different view has a different importance or contribution during the clustering processes. Thus, we formulate the objective function of multi-view k-means clustering (MVKMC) as below.

\begin{equation}
J_{\text{MVKMC}} (V, U, A) = \sum_{h=1}^s v_h^{\alpha} \sum_{i=1}^n \sum_{k=1}^c \mu_{ik} \|x_i^h - a_k^h\|^2 
\label{eqn:MVKMC}
\end{equation}

\begin{equation*}
\text{s.t.} \sum_{k=1}^c \mu_{ik} = 1, \mu_{ik} \in \{0,1\} \text{~~and~~} \sum_{h=1}^s v_h = 1, v_h \in [0,1]
\end{equation*}

where $\alpha$ is an exponent parameter to control the behavior of one data view during the clustering process. In our framework, we avoid having a singularity matrix or sparse values within one component of view’ vector during the iteration process. Thus, we are not recommended $\alpha<0$.  When $\alpha=0$,  $J_{\text{MVKMC}} (V,U,A)$ is treated all data views equally importance. Here, we must assign $\alpha>1$. It is worth to note that the $\alpha$ parameter throughout experiments in this paper initialized to be 2, as if they are updated the model will be able to learn and converge faster. In that sense, $\alpha=2$ is recognized as an advanced number and adjustable on the output during the clustering process of MVKMC.

\subsection*{The Optimization}

\begin{equation}
\mu_{ik}^h=
\begin{cases}
1      &\text{if}\ \sum_{h=1}^s  (v_h)^{\alpha} \|x_i^h - a_k^h \|^2  = \min\limits_{1\leq k\leq c} \sum_{h=1}^s  (v_h)^{\alpha} \|x_i^h - a_k^h \|^2 \\
0      & \text{Otherwise}
\end{cases}
\label{eqn:updateU}
\end{equation}

\begin{equation}
v_h = \sum_{h^{\prime} = 1}^s \bigg ( \frac{1 / \sum_{i=1}^n \sum_{k=1}^c \mu_{ik}\|x_i^h - a_k^h\|^2} {1 / \sum_{i=1}^n \sum_{k=1}^c \mu_{ik}\|x_i^{h^{\prime}} - a_k^{h^{\prime}}\|^2} \bigg)^{\frac{1} {\alpha - 1}}
\label{eqn:updateV}
\end{equation}

\begin{equation}
a_{kj}^h = \frac{v_h^2 \sum_{i=1}^n \mu_{ik}}{v_h^2 \sum_{i=1}^n \mu_{ik}} x_{ij}^h
\label{eqn:updateA}
\end{equation}

The whole procedure for solving problem \ref{eqn:MVKMC} is summarized in Algorithm \ref{alg:mvkmc}.

\begin{algorithm}
\caption{The MVKMC}
\label{alg:mvkmc}
\begin{algorithmic}
\Require $X = \{x_1, x_2, \ldots, x_n\}$ with $x_i = \{x_i^h\}_{h=1}^s $ and $x_i^h = \{x_{ij}^h\}_{j=1}^{d_h}$, the number of clusters $c$, the number of views $s$, $t=0$, and $\varepsilon > 0$.
\Ensure $A^{h(0)} = [a_{kj}^h]_{c \times d_h}$ and $V^{(0)} = [v_h]_{1 \times s}$ with $v_h = \frac{1}{s} ~\forall h$
\While{not converged }
\State Compute the memberships $\mu_{ik}$ by using Eq. \ref{eqn:updateU}
\For{$h=1 \text{~to~} s$}
\State Update the weights $v_h$ by using Eq. \ref{eqn:updateV}
\State Update the cluster centers $a_{kj}^j$ by using Eq. \ref{eqn:updateA}
\EndFor
\State Check the condition of convergence $\|A^{h^{(t)}} - A^{h^{(t-1)}}\| < \varepsilon$
\EndWhile
\State Apply MVKMC algorithm to encoding the labels
\State \textbf{Output: }Clustering labels for input data X
\end{algorithmic}

\end{algorithm}

\subsection{Multi-view k-means with exponent distance}

Kernel learning based weighted distance between data points and cluster centers of different views from multi-view data is effortless and required to quantify various uncertainties behavior during approximation processes. Especially uncertainty data sources with high sparsity and noise. Motivated by these issues, we are projecting the Euclidean norm $\|x_i^h -a_k^h\|^2$ by adopting the nonlinear structure of hierarchical Bayesian to develop our proposed multi-view k-means clustering algorithm that leads to the likelihood posterior distribution such as  $-\text{exp} (-\beta^h \|x_i^h -a_k^h\|^2)$. Using the representation of Bayesian algorithm, a very different approach of metric to group comparable data points into the same cluster based on a smooth movement, we further extended Eq. \ref{eqn:MVKMC} in the following way.

\subsection*{The Objective Function}

\begin{equation}
J_{\text{MVKM-ED}} (V,U,A) = \sum_{h=1}^s v_h^{\alpha} \sum_{i=1}^n \sum_{k=1}^c \mu_{ik} \bigg{\{} 1- \text{exp} \bigg ( - \beta^h \|x_i^h -a_k^h\|^2\bigg) \bigg{\}}
\label{eqn:mvkmed}
\end{equation}

\begin{equation*}
\text{s.t.} \sum_{k=1}^c \mu_{ik} = 1, \mu_{ik} \in \{0,1\} \text{~~and~~} \sum_{h=1}^s v_h = 1, v_h \in [0,1]
\end{equation*}

where  $\alpha$ is an exponent parameter to control the distribution of view weights and $\beta^h$ is the corresponding kernel coefficients to regularize the distance between data points  $x_{ij}^h$ and cluster centers  $a_{kj}^h$. For all the experiments throughout the paper, we assign $\alpha\geq 2$ and estimate  $\beta^h$ by the following equations. 

\begin{equation}
\beta^h = \frac{c}{tn} \sum_{j=1}^{d_h} \bar{x}_{ij}^{d_h}
\label{eqn:betaest1}
\end{equation}

\begin{equation}
\beta^h = \frac{\sum_{i=1}^n \| x_i^h - \bar{x}_i^h\|}{n}
\label{eqn:betaest2}
\end{equation}

\begin{equation}
\beta^h =  \max\limits_{1\leq k\leq c} \bigg( \sqrt{\frac{\sum_{i=1}^n \| x_i^h - a_k^h\|}{n}} \bigg) - \min\limits_{1\leq k\leq c} \bigg( \sqrt{\frac{\sum_{i=1}^n \| x_i^h - a_k^h\|}{n}} \bigg) 
\label{eqn:betaest3}
\end{equation}

Where $\bar{x}_{ij}^h = \sum_{j=1}^{d_h} x_{ij} / n$ is used to identify the effect of a central tendency in handling the outliers on MV data. Here, the associated estimators are doubly robust to exposure the characteristics of MV data. The expected value of  $\beta^h$ are distributed under intervals of  $[0,1]$   or $>2$.

\subsection*{The Optimization}
Optimizing  $J_{\text{MVKM-ED}} (V,U,A)$ with respect to  $\mu_{ik}$  is equivalent to minimizing the problem $\tilde{J}_{\text{MVKM-ED}} (V,U,A, \lambda_1) = \text{arg} \min\limits_{1\leq k\leq c} \sum_{h=1}^s v_h^{\alpha} \sum_{i=1}^n \sum_{k=1}^c \mu_{ik} \bigg{\{} 1- \text{exp} \bigg ( - \beta^h \|x_i^h -a_k^h\|^2\bigg) \bigg{\}} - \lambda_1 \sum_{k=1}^c (\mu_{ik} -1)$. The partial derivation of  $\tilde{J}_{\text{MVKM-ED}} (V,U,A, \lambda_1) $ with respect to $\mu_{ik}$  can be expressed as $\partial ( \text{arg} \min\limits_{1\leq k\leq c} \sum_{h=1}^s v_h^{\alpha} \sum_{i=1}^n \sum_{k=1}^c \mu_{ik} \bigg{\{} 1- \text{exp} \bigg ( - \beta^h \|x_i^h -a_k^h\|^2\bigg) \bigg{\}} - \lambda_1 \sum_{k=1}^c (\mu_{ik} -1) ) / \partial \mu_{ik} = 0$ . We obtain that  . Since  , we can formulate the optimal solution for sub-problem of  $\mu_{ik}$ in the following way.

\begin{equation}
\mu_{ik}=
\begin{cases}
1      &\text{if}\ \sum_{h=1}^s  (v_h)^{\alpha}  \bigg{\{} 1- \text{exp} \bigg ( - \beta^h d_{ik}^h \bigg) \bigg{\}}  = \min\limits_{1\leq k\leq c} \sum_{h=1}^s  (v_h)^{\alpha}  \bigg{\{} 1- \text{exp} \bigg ( - \beta^h d_{ik}^h \bigg) \bigg{\}} \\
0      & \text{Otherwise}
\end{cases}
\label{eqn:updateU2}
\end{equation}

where $d_{ik}^h = \| x_i^h - a_k^h\|$. Next, to find the partial derivative of  $J_{\text{MVKM-ED}} (V,U,A)$  with respect to  $a_{kj}^h$, we need to consider the terms that depend on  $a_{kj}^h$ and differentiate them while treating all other variables as constant, such as $\partial J_{\text{MVKM-ED}} (V,U,A) / \partial a_{kj}^h  = -2 v_h^{\alpha} \sum_{i=1}^n \mu_{ik} (x_{ij}^h - a_{kj}^h) \text{ exp} (-\beta^h d_{ik}^h) = 0 $. We get $v_h^{\alpha} \sum_{i=1}^n \mu_{ik} \text{ exp} (-\beta^h d_{ik}^h) a_{kj}^h = v_h^{\alpha} \sum_{i=1}^n \mu_{ik} \text{ exp} (-\beta^h d_{ik}^h) x_{ij}^h$ and $v_h^{\alpha} \sum_{i=1}^n \mu_{ik} \varphi_{ik}^h a_{kj}^h = v_h^{\alpha} \sum_{i=1}^n \mu_{ik} \varphi_{ik}^h x_{ij}^h$, where $\varphi_{ik}^h = \text{ exp} (-\beta^h d_{ik}^h)$. Therefore, the updating equation of  $J_{\text{MVKM-ED}} (V,U,A)$   with respect to  $a_{kj}^h$ can be expressed in the following way.

\begin{equation}
a_{kj}^h = \frac{v_h^{\alpha} \sum_{i=1}^n \mu_{ik} \varphi_{ik}^h} {v_h^{\alpha}  \sum_{i=1}^n \mu_{ik} \varphi_{ik}^h} x_{ij}^h
\label{eqn:updateA2}
\end{equation}

The Lagrange of  $J_{MVKM-ED} (V,U,A)$  with respect to $v_h$ can be expressed as $\tilde{J}_{\text{MVKM-ED}} (V,U,A, \lambda_2) = \text{arg} \min\limits_{1\leq k\leq c} \sum_{h=1}^s v_h^{\alpha} \sum_{i=1}^n \sum_{k=1}^c \mu_{ik} \bigg{\{} 1- \text{exp} \bigg ( - \beta^h \|x_i^h -a_k^h\|^2\bigg) \bigg{\}} - \lambda_2 \sum_{h=1}^s (v_h -1)$. By taking the first partial derivative of $\tilde{J}_{\text{MVKM-ED}} (V,U,A, \lambda_2)$ with respect to $v_h$, we have $\alpha v_h^{\alpha -1} \sum_{i=1}^n \sum_{k=1}^c \mu_{ik}   \bigg{\{} 1- \text{exp} \bigg ( - \beta^h \varphi_{ik}^h \bigg) \bigg{\}} - \lambda_2 = 0$ and $v_h = (\lambda_2)^{\frac{1}{\alpha - 1}} \bigg ( 1 /  \sum_{i=1}^n \sum_{k=1}^c \mu_{ik}   \bigg{\{} 1- \text{exp} \bigg ( - \beta^h \varphi_{ik}^h \bigg) \bigg{\}}  \bigg )^{1/\alpha -1}$. Since $\sum_{h=1}^s v_h =1$, we get $\lambda_2 = 1 /  \sum_{h^{\prime}=1}^s \bigg (  \sum_{i=1}^n \sum_{k=1}^c \mu_{ik}   \bigg{\{} 1+ \eta^{h^{\prime}} \bigg) \bigg{\}} \bigg )^{-1/\alpha -1}$ with $\eta^h =  - \text{ exp} (-\beta^h d_{ik}^h) $. Thus, we have the update function for view weight $v_h$ as follows.

\begin{equation}
v_h = \sum_{h^{\prime}=1}^s \bigg ( \frac{ \sum_{i=1}^n \sum_{k=1}^c \mu_{ik} \{ 1 + \eta^{h^{\prime}} \} } {\sum_{i=1}^n \sum_{k=1}^c \mu_{ik} \{ 1 + \eta^h \} } \bigg )^{-\frac{1}{\alpha -1}}
\label{eqn:updateV2}
\end{equation}

The whole procedure for solving problem \ref{eqn:mvkmed} is summarized in Algorithm \ref{alg:mvkmc-ed}.

\begin{algorithm}
\caption{The MVKM-ED}
\label{alg:mvkmc-ed}
\begin{algorithmic}
\Require $X = \{x_1, x_2, \ldots, x_n\}$ with $x_i = \{x_i^h\}_{h=1}^s $ and $x_i^h = \{x_{ij}^h\}_{j=1}^{d_h}$, the number of clusters $c$, the number of views $s$, $\beta^h$, $t=0$, and $\varepsilon > 0$.
\Ensure $A^{h(0)} = [a_{kj}^h]_{c \times d_h}$ and $V^{(0)} = [v_h]_{1 \times s}$ with $v_h = \frac{1}{s} ~\forall h$
\While{not converged }
\State Compute the memberships $\mu_{ik}$ by using Eq. \ref{eqn:updateU2}
\For{$h=1 \text{~to~} s$}
\State Estimate the kernel coefficients $\beta^h$ by Eqs. \ref{eqn:betaest1}, \ref{eqn:betaest2} or \ref{eqn:betaest3}
\State Update the weights $v_h$ by using Eq. \ref{eqn:updateV2}
\State Update the cluster centers $a_{kj}^j$ by using Eq. \ref{eqn:updateA2}
\EndFor
\State Check the condition of convergence $\|J_{\text{MVKM-ED}}^{h^{(t)}} - J_{\text{MVKM-ED}}^{h^{(t-1)}}\| < \varepsilon$
\EndWhile
\State Apply MVKM-ED or k-means algorithm to encoding the labels
\State \textbf{Output: }Clustering labels for input data X
\end{algorithmic}

\end{algorithm}

\subsection{The Gaussian-kernel multi-view k-means} 

\subsection*{The Objective Function}

A new investigation for a new unsupervised manner to recognize multi-view pattern data with kernel-distance-based approach with Euclidean norm minimization is representing in this sub-section. We will learn the potential of  $p$ stabilizer parameter on MVKM-ED for solving the multi-view data. To penalize the effects of $p$, a new condition in the weighted distance $-\text{exp}(-\beta^h \|x_i^h - a_k^h\|^2)$ is regularized as $(-\text{exp}(-\beta^h \|x_i^h - a_k^h\|^2))^p$. Following this new regularization, the corresponding objective function in Eq. \ref{eqn:mvkmed} is modified in the following way.

\begin{equation}
J_{\text{GKMVKM}} (V,U,A) = \sum_{h=1}^s v_h^{\alpha} \sum_{i=1}^n \sum_{k=1}^c \mu_{ik} \bigg{\{} 1- \bigg( \text{exp} \bigg ( - \beta^h \|x_i^h -a_k^h\|^2\bigg) \bigg)^p\bigg{\}}
\label{eqn:gkmvkm}
\end{equation}

\begin{equation*}
\text{s.t.} \sum_{k=1}^c \mu_{ik} = 1, \mu_{ik} \in \{0,1\} \text{~~and~~} \sum_{h=1}^s v_h = 1, v_h \in [0,1]
\end{equation*}

where  $\beta^h = \bigg( n / \sum_{i=1}^n \|x_i^h - \bar{x}_i^h \|^2\bigg)$,  $\bar{x}_i^h = \sum_{i=1}^n x_i^h / n$, and a stabilizer $p$  are simultaneously used to reduce the effect of insignificant or noises data during clustering processes. Addittionaly, in GKMVKM, these parameters of $\beta^h$  and $p$ are extensively playing major roles and should be carefully tuned to the problem at hand. 

\subsection*{The Optimization}

To enable the iterative processes in  $J_{\text{GKMVKM}} $, following the updating function that can be uses to solve these sub-problems of  $\mu_{ik}$, $v_h$, and  $a_{kj}^h$. 

\begin{equation}
\mu_{ik}=
\begin{cases}
1      &\text{if}\ \sum_{h=1}^s  (v_h)^{\alpha}  \bigg{\{} 1- \bigg(\text{exp} \bigg ( - \beta^h d_{ik}^h \bigg)\bigg)^p \bigg{\}}  = \min\limits_{1\leq k\leq c} \sum_{h=1}^s  (v_h)^{\alpha}  \bigg{\{} 1- \bigg( \text{exp} \bigg ( - \beta^h d_{ik}^h \bigg)\bigg)^p \bigg{\}} \\
0      & \text{Otherwise}
\end{cases}
\label{eqn:updateU3}
\end{equation}

\begin{equation}
a_{kj}^h = \frac{v_h^{\alpha} \sum_{i=1}^n \mu_{ik} \rho_{ik}^h} {v_h^{\alpha}  \sum_{i=1}^n \mu_{ik} \rho_{ik}^h} x_{ij}^h
\label{eqn:updateA3}
\end{equation}

\begin{equation}
v_h = \sum_{h^{\prime}=1}^s \bigg ( \frac{ \sum_{i=1}^n \sum_{k=1}^c \mu_{ik} \{ 1 - \rho^{h^{\prime}} \} } {\sum_{i=1}^n \sum_{k=1}^c \mu_{ik} \{ 1 - \rho^h \} } \bigg )^{-\frac{1}{\alpha -1}}
\label{eqn:updateV3}
\end{equation}

where $\rho_{ik}^h = \bigg(\text{exp} \bigg ( - \beta^h d_{ik}^h \bigg)\bigg)^p$.

\subsection*{A procedure to estimate a stabilizer $p$}

To tune a stabilizer $p$ in GKMVKM, we proposed two procedures such as user defined parameter and using estimator function. For user defined parameter, we recommend $p\geq2$. While for estimator function, we used the concept of peak mountain function of data points  $M(X)$. The peak mountain function is computed by an exponent weighted of sum of the squared distances between $x_i^h$  and $a_k^h$ in different views with addition to an exponent weighted of sum of the squared distances between means  $x_i^h$ and means of  $\bar{x}_{[c]i}^h$ in different views. To get a single value as a peak mountain, we take the minimum of maximum value of those two additions in different views considering its partition of multiple classes. In that sense, to enable a mountain function estimation, we must create a partition stage based on the representation of multiple class across multiple features data, independently. Our mountain function is formulated as below.

\begin{equation}
\text{Opt}(p) = M(X) = \min\limits_{1\leq k\leq c} \bigg [ \bigg (  s/ \max\limits_{1\leq k\leq c}  \sum_{i=1}^n \sum_{k=1}^c  \text{exp} \bigg ( - \beta^h \|x_i^h -a_k^h\|^2\bigg)  + \text{exp} \bigg ( - \beta^h \|\bar{x}_i^h -\bar{x}_{[c]j}^h\|^2\bigg)  \bigg ] \bigg )
\label{eqn:optp}
\end{equation}

where $\bar{x}_i^h = \frac{1}{n} \sum_{i=1}^n x_{ij}^h$  and $\bar{x}_{[c]i}^h = \frac{1} {n_i \in c_k} \sum_{i \in c_k} x_{ij}^h$. It is worth to note that the procedure for selecting the optimal p stabilizer parameter using the concept of peak mountain as presented in Eq. \ref{eqn:optp} can be extremely expensive as it is involves a partition stage of multiple classes based on the representation of multiple features data, the exponential function in Eq. \ref{eqn:optp} is computationally heavy and slow. Different assignments for initial number of instances belong to one class may effect the performance of clustering results as it is distribute the data points into different lists of multiple classes. In that sense, the good estimator for an optimal stabilizer parameter p is still questionable. User define procedure may the best one as each user is to be assumed know their data and what they want to discover. The whole procedure for solving problem \ref{eqn:gkmvkm} is summarized in Algorithm \ref{alg:GKMVKM}.

\begin{algorithm}
\caption{The GKMVKM}
\label{alg:GKMVKM}
\begin{algorithmic}
\Require $X = \{x_1, x_2, \ldots, x_n\}$ with $x_i = \{x_i^h\}_{h=1}^s $ and $x_i^h = \{x_{ij}^h\}_{j=1}^{d_h}$, the number of clusters $c$, the number of views $s$, the initial stabilizer $p$ is defined by users or  can be estimated by Eq. \ref{eqn:optp}, $t=0$, and $\varepsilon > 0$.
\Ensure $A^{h(0)} = [a_{kj}^h]_{c \times d_h}$ and $V^{(0)} = [v_h]_{1 \times s}$ with $v_h = \frac{1}{s} ~\forall h$
\While{not converged }
\For{$h=1 \text{~to~} s$}
\State Estimate the kernel coefficients $\beta^h$
\State Compute the memberships $\mu_{ik}$ by using Eq. \ref{eqn:updateU3}
\State Update the cluster centers $a_{kj}^j$ by using Eq. \ref{eqn:updateA3}
\State Update the weights $v_h$ by using Eq. \ref{eqn:updateV3}
\EndFor
\State Check the condition of convergence $\|J_{\text{GKMVKM}}^{h^{(t)}} - J_{\text{GKMVKM}}^{h^{(t-1)}}\| < \varepsilon$
\EndWhile
\State Apply GKMVKM algorithm to encoding the labels
\State \textbf{Output: }Clustering labels for input data X
\end{algorithmic}

\end{algorithm}

\section{Experiments}
\subsection{Data set and evaluation measures}

\subsubsection{Real-world data}

In our experiment, five real-world multi-view data are used, i.e., UWA3D \cite{wang2016adaptive}, Newsgroups (NGs) \footnote{http://qwone.com/~jason/20Newsgroups/}, hundred plant species leaves \cite{misc_one-hundred_plant_species_leaves_data_set_241}, ORL-Face  \footnote{https://www.v7labs.com/open-datasets/orl}, and NUS-WIDE object data \cite{chua2009nus}. Capturing motion or human motion retrieval is a challenging task in computer animation and multimedia analysis communities \cite{wang2016adaptive}. Some necessary category and multi-view information about five datasets are summarized in Table~\ref{tab:table1}, where $c$, $s$, and $d_h$ represent the number of clusters, views, and dimensions, respectively. The UWA3D multi-view activity contains 660 action sequences that 11 actions performed by 12 subjects with five repetitive actions. The UWA3D data extracted with RGB and depth features. Newsgroups (NGs) stands for one of the subsets of 20 Newsgroups data. The 20 Newsgroups (20NGs) data has 500 instances and describing by 2000 features from three different views. The hundred plant species leaves dataset consists of 16 images from 100 different plants. For these 1600 leaves are described by three views such as margin, shape, and texture \cite{beghin2010shape}. The ORL-Face data contains of 400 images from 40 distinct subjects. These 400 images were featuring the facial expressions, facial details, and taken by varying the lighting. The NUS-WIDE Object data is mostly used for object recognition purposes. This data originally collected by Chua et al. \cite{chua2009nus} with 269,648 images and a total of 5,018 tags collected from Flickr. 

\begin{table}
 \caption{The Summary of Five Real-World Applications of Multi-View Dataset}
  \centering
  \begin{tabular}{lllll}
    \toprule
    Dataset     & Number of $c$     & Number of $s$ & Number of $n$ & Number  of $d_h$ \\
    \midrule
     UWA3D & 30 & 2 & 254 & $[6144~110]$   \\
     NGs     & 20 & 3 & 500 & $[2000~2000~2000]$    \\
     100 Leaves     & 100  & 3& 1600 & $[64~64~64]$ \\
     ORL-Face & 40 & 4 & 400 & $[4096 ~ 3304~ 6750 ~ 1024]$\\
     NUS-WIDE & 12 & 6 & 2400 & $[64~144~73~128~225~500]$\\
    \bottomrule
  \end{tabular}
  \label{tab:table1}
\end{table}

\subsubsection{Synthetic data}

\begin{figure}[!b]
      \centering
      \subfloat[View 1]{\includegraphics[width=.3\textwidth]{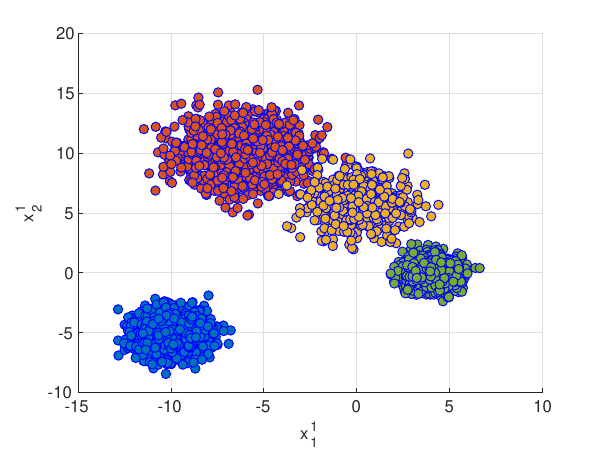}}
      \qquad
      \subfloat[View 2]{\includegraphics[width=.3\textwidth]{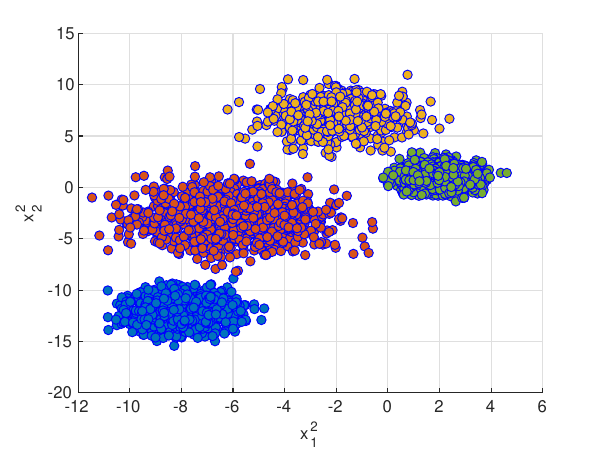}}
      \qquad
      \subfloat[View 3]{\includegraphics[width=.3\textwidth]{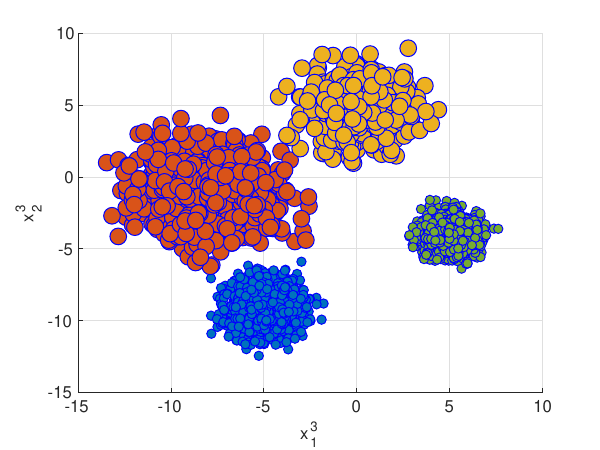}}
  \caption{The visualization of 3-views-4-clusters dataset}
  \label{fig:fig1}
\end{figure}

10000 instances of three-view numerical data set with 4 clusters and 2 feature components are considered. The data points in each view are generated from a 2-component 2-variate Gaussian mixture model (GMM) where their mixing proportions $\alpha_1^{(1)}=\alpha_1^{(2)}=\alpha_1^{(3)}=\alpha_1^{(4)}=0.3$;  $\alpha_2^{(1)}=\alpha_2^{(2)}=\alpha_2^{(3)}=\alpha_2^{(4)}=0.15$; $\alpha_3^{(1)}=\alpha_3^{(2)}=\alpha_3^{(3)}=\alpha_3^{(4)}=0.15$  and $\alpha_4^{(1)}=\alpha_4^{(2)}=\alpha_4^{(3)}=\alpha_4^{(4)}=0.4$. The means  $\mu_{ik}^{(1)}$ for the first view are $[-10 ~-5)]$,$[-9 ~ 11]$, $[0~ 6]$   and $[4~0]$;  The means  $\mu_{ik}^{(2)}$ for the view 2 are $[-8 ~-12]$,$[-6 ~ -3]$, $[-2~ 7]$   and $[2~1]$; And the means  $\mu_{ik}^{(3)}$ for the third view are $[-5 ~-10]$,$[-8 ~ -1]$, $[0~ 5]$   and $[5~-4]$. The covariance matrices for the three views are $\Sigma_1^{(1)}=\Sigma_1^{(2)}=\Sigma_1^{(3)}=\Sigma_1^{(4)}=\left[ \begin{array}{cc} 1 & 0\\0&1\end{array}\right]$; $\Sigma_2^{(1)}=\Sigma_2^{(2)}=\Sigma_2^{(3)}=\Sigma_2^{(4)}=3 \left[ \begin{array}{cc} 1 & 0\\0&1\end{array}\right]$;  $\Sigma_3^{(1)}=\Sigma_3^{(2)}=\Sigma_3^{(3)}=\Sigma_3^{(4)}=2 \left[ \begin{array}{cc} 1 & 0\\0&1\end{array}\right]$; and $\Sigma_4^{(1)}=\Sigma_4^{(2)}=\Sigma_4^{(3)}=\Sigma_4^{(4)}=0.5 \left[ \begin{array}{cc} 1 & 0\\0&1\end{array}\right]$. These $x_1^{(1)}$  and $x_2^{(1)}$  are the coordinates for the view 1,  $x_1^{(2)}$  and  $x_2^{(2)}$ are the coordinates for the view 2,  $x_1^{(3)}$ and $x_2^{(3)}$ are the coordinates for the view 3. While the original distribution of data points for cluster 1, cluster 2, cluster 3, and cluster 4 are 1514, 3046, 3903, and 1537, respectively. Figure \ref{fig:fig1} displays this 3-views-4-clusters data.

\subsubsection{Evaluation measures}

To quantify the performance of our proposed MVKMC, MVKM-ED, and GKMVKM, we make comparison against these five related methods of RMKMC \cite{cai2013multi}, \cite{xu2016weighted}, SWVF \cite{jiang2016multi}, TW-k-means \cite{chen2011tw}, and Tw-Co-k-means \cite{zhang2018tw} on the five benchmark multi-view data. For all these algorithms, the minimum, average and maximum value of its normalized mutual information (NMIs), adjusted rand index (ARIs), accuracy (ACC), recall, precision, and F-score \cite{hubert1985comparing, manning2008introduction} are reported based on their 50 different initializations. Note that large values of NMI, ARI, ACC, recall, precision, and F-score indicate better clustering performance. For performance comparisons, the highest performance is in boldface, the second-best performance is in underlined, and the third-best performance is in italic. 

\subsection{Clustering Results}

\subsection*{Result 1: MVKMC and MVKM-ED on NGs Dataset}

We first conducted experiments on NGs data to measure the performance of the proposed MVKMC and MVKM-ED. For the comparison algorithms, we set $\alpha=0.000003$, $\beta=0.0025$ to WMCFS, $\alpha=8$, $\beta=0.0025$ to SWVF, $\eta=10$, $\beta=5$ to TW-k-means, $\alpha=60$, $\beta=50$, $\eta=0.45$ to TW-Co-k-means. We implemented the exponent parameter $\alpha=2$ to MVKMC and $\alpha=9$ to MVKM-ED and the result are reported in Table~\ref{tab:result1}. Reading from Table~\ref{tab:result1}, we notice that the proposed MVKM-ED is hugely robust as compared to the other methods. The result shown that the exponential-distance-based approach can obtain a better estimate and more accurate clustering performances showing its potential to discover multi-view patterns data.

\begin{table}
\tiny
 \caption{The Clustering Performances on NGs Dataset}
  \centering
  \begin{tabular}{lllllll}
  \\
    \toprule
         & NMI & ARI & ACC & Recall & Precision & F-Score \\
    \midrule
    RMKMC&	\underline{0.4707}/\underline{0.7609}/\textit{0.8904}	&\underline{0.3500}/\underline{0.7438}/\emph{0.9064}&	\underline{0.6020}/\underline{0.8579}/\emph{0.9620}	&\emph{0.5079}/\emph{0.8080}/0.9270&	\underline{0.4638}/\underline{0.7853}/\emph{0.9230}&	\underline{0.4848}/\underline{0.7960}/\emph{0.9250}\\
   WMCFS &	0.1813/0.2781/0.3715&	0.1073/0.1904/0.2691&	0.4060/0.5011/0.5900	&0.3358/0.4235/0.4861	&0.2654/0.3310/0.3899&	0.3067/0.3708/0.4284\\
   SWVF&	0.3592/0.6451/0.8610&	0.1190/0.5719/0.8871&	0.3620/0.7350/0.9540&	0.5248/0.7438/0.9108	&0.2542/0.6194/0.9082&	0.3811/0.6688/0.9095\\
   TW-k-means&	0.0155/0.0510/0.2285&	0.0000/0.0079/0.1756&	0.2020/0.2263/0.3660&	\underline{0.7000}/\underline{0.9360}/\textbf{0.9842}&	0.1984/0.2022/0.2874&	0.3205/0.3315/0.4075\\
   TW-Co-k-means&	\emph{0.4283}/0.6218/0.8386&	\emph{0.2843}/0.5074/0.8267&	\emph{0.4960}/0.7042/0.9300&	0.4905/0.7106/0.8724&	0.3517/0.5573/0.8508&	\emph{0.4768}/0.6198/0.8615\\
   MVKMC& 	0.4006/\emph{0.7332}/\underline{0.9303}&	0.2677/\emph{0.6858}/\underline{0.9454}&	0.4580/\emph{0.8030}/\underline{0.9780}&	0.4769/0.7943/\emph{0.9564}&	\emph{0.3880}/\emph{0.7195}/\underline{0.9560}&	0.4279/\emph{0.7530}/\underline{0.9562}\\
   MVKM-ED&	\textbf{0.9001}/\textbf{0.9142}/\textbf{0.9355}&	\textbf{0.9155}/\textbf{0.9305}/\textbf{0.9501}&	\textbf{0.9660}/\textbf{0.9721}/\textbf{0.9800}&	\textbf{0.9337}/\textbf{0.9453}/\underline{0.9605}&	\textbf{0.9307}/\textbf{0.9433}/\textbf{0.9595}&	\textbf{0.9323}/\textbf{0.9443}/\textbf{0.9600}\\
    \bottomrule
  \end{tabular}
  \label{tab:result1}
\end{table}

\subsection*{Result 2: Robustness to  a large number of observations}

To evaluate the ability of our proposed GKMVKM to perform well under a large number of data instances, we conduct experiments on synthetic and NUS-WIDE object data. On synthetic data, we set  $\alpha=4$, $p=3$ to GKMVKM, $\alpha=25$, $\beta=0.1$ to WMCFS, $\alpha=8$, $\beta=0.0025$ to SWVF, $\eta=10$, $\beta=5$ to TW-k-means, $\alpha=60$, $\beta=50$, $\eta=0.45$ to TW-Co-k-means. While to see the effect of large number of data instances on NUS-WIDE, we let $\alpha=4$, $p=2$ to GKMVKM, $\alpha=12$, $\beta=0.0000057$ to WMCFS, $\alpha=8$, $\beta=0.0025$ to SWVF, $\eta=10$, $\beta=5$ to TW-k-means, $\alpha=60$, $\beta=50$, $\eta=0.45$ to TW-Co-k-means. To provide a fair comparison, we initialized the cluster centers and memberships of each algorithm uses single-view k-means clustering over 50 times runs. Tables~\ref{tab:result2}-\ref{tab:result3} reports these minimum, average, and maximum scores of NMI, ARI, ACC, Recall, Precision, and F scores performed by our proposed techniques (MVKMC, MVKM-ED, GKMVKM) and five related methods on synthetic and NUS-WIDE Object datasets, respectively. As shown in Table~\ref{tab:result2}, the highest NMI, ARI, ACC, Recall, Precision, and F Score are achieved by either MVKM-ED or GKMVKM. The second highest scores are obtained by the Tw-k-means. And the third highest scores are obtained by Tw-Co-k-means. Furthermore, the experimental results on NUS-WIDE data also show an agreement that the proposed GKMVKM can achieve a promising result and gather accurate highest minimum NMI, average and maximum recall. All the information we can verify from these experiments is the proposed GKMVKM works quite well when the number of observations or instances is large.

\begin{table}
\tiny
 \caption{The Clustering Performances on Synthetic Dataset}
  \centering
  \begin{tabular}{lcccccc}
    \toprule
         & NMI & ARI & ACC & Recall & Precision & F-Score \\
    \midrule
    RMKMC&	0/0/0	&0/0/0&	-&	0/0/0&	0.2916/0.2916/0.2916&	0.4515/0.4515/0.4515\\
    WMCFS&	0/0.5013/\textbf{1}&	0/0.4919/	\textbf{1}&	0.3903/0.6856/	\textbf{1}&	\underline{0.9450}/\underline{0.9956}/	\textbf{1}&	0.2916/0.6307/	\textbf{1}& 0.4515/0.7180/	\textbf{1}\\
    SWVF	&\underline{0.7826}/0.9426/\emph{0.9988}&	0.6286/0.9031/0.9995&	0.6964/0.9214/\emph{0.9998}&	\emph{0.8399}/0.9582/\emph{0.9998}&	0.6753/0.9152/0.9995&	\emph{0.7487}/0.9344/0.9996\\
    TW-k-means	&0.7643/\underline{0.9599}/	\textbf{1}&	\underline{0.6302}/\underline{0.9470}/	\textbf{1}&	\emph{0.6966}/\underline{0.9504}/	\textbf{1}&	0.8226/\textit{0.9703}/	\textbf{1}	&\underline{0.6808}/\underline{0.9576}/	\textbf{1}	&\underline{0.7488}/\underline{0.9633}/	\textbf{1}\\
    TW-Co-k-means&	\emph{0.7769}/\emph{0.9468}/\underline{0.9993}&	\emph{0.6289}/\emph{0.9216}/\underline{0.9997}&	\underline{0.6967}/\emph{0.9281}/\underline{0.9999}&	0.8210/0.9594/\underline{0.9999}	&\emph{0.6766}/\emph{0.9353}/\underline{0.9997}&	0.7486/\emph{0.9461}/\underline{0.9998}\\
    MVKMC&	0.7695/0.9411/\emph{0.9988}&	0.6276/0.9111/\emph{0.9996}&	0.6959/0.9228/\emph{0.9998}&	0.8281/0.9568/\emph{0.9998}&	0.6763/0.9252/\emph{0.9996}&	0.7477/0.9393/\emph{0.9997}\\
    MVKM-ED	&\textbf{1/1/1}&	\textbf{1/1/1}	&\textbf{1/1/1}	&\textbf{1/1/1}	&\textbf{1/1/1}&	\textbf{1/1/1}\\
    GKMVKM&	\textbf{1/1/1}&	\textbf{1/1/1}&	\textbf{1/1/1}	&\textbf{1/1/1}	&\textbf{1/1/1}&	\textbf{1/1/1}\\
    \bottomrule
  \end{tabular}
  \label{tab:result2}
\end{table}

\begin{table}
\tiny
 \caption{The Clustering Performances on NUS-WIDE Object Dataset}
  \centering
  \begin{tabular}{lcccccc}
    \toprule
         & NMI & ARI & ACC & Recall & Precision & F-Score \\
    \midrule
  	  RMKMC	&0.1125/\textbf{0.1281}/\textbf{0.1444}&	\textit{0.0478}/\textbf{0.0640}/\textbf{0.0828}&	0.2104/\textbf{0.2380}/\textbf{0.2850}&	0.1378/0.1531/0.1709&	\textbf{0.1244}/\textbf{0.1386}/\textbf{0.1564}&	0.1311/\emph{0.1455}/\textbf{0.1626}\\
	WMCFS	&\underline{0.1156}/\emph{0.1257}/0.1347&	\textbf{0.0512}/\underline{0.0591}/0.0696&	\emph{0.2113}/\underline{0.2366}/0.2563&	\textbf{0.1562}/\underline{0.1679}/0.1844&	\underline{0.1229}/\underline{0.1307}/0.1414&	\textbf{0.1413}/\textbf{0.1468}/\emph{0.1558}\\
	SWVF	&0.1013/0.1151/0.1260&	0.0459/0.0530/0.0632&	0.2000/0.2214/0.2467&	0.1398/\emph{0.1650}/0.1852&	0.1189/0.1252/0.1343&	0.1308/0.1423/0.1500\\
	TW-k-means	&0.0856/0.0967/0.1046&	0.0326/0.0411/0.0456&	0.1746/0.1993/0.2183&	0.1445/0.1615/\emph{0.1885}&	0.1079/0.1148/0.1195&	0.1272/0.1340/0.1412\\
	TW-Co-k-means	&\emph{0.1100}/0.1234/0.1363&	0.0467/\emph{0.0573}/\emph{0.0703}&	\textbf{0.2125}/0.2309/0.2592&	\emph{0.1485}/0.1621/0.1811&	\emph{0.1207}/\emph{0.1299}/\emph{0.1424}&	\emph{0.1354}/0.1441/0.1534\\
	MVKM	&0.1043/0.1200/\textit{0.1356}&	0.0418/0.0525/\underline{0.0742}&	0.1971/0.2219/\underline{0.2642}&	0.1477/0.1647/\underline{0.1893}&	0.1160/0.1249/\underline{0.1429}&	0.1327/0.1419/\underline{0.1603}\\
	GKMVKM	&\textbf{0.1159}/\underline{0.1279}/\underline{0.1365}&	\underline{0.0480}/0.0544/0.0623&	\underline{0.2117}/\emph{0.2326}/\emph{0.2600}&	\underline{0.1527}/\textbf{0.1731}/\textbf{0.2070}&	0.1190/0.1256/0.1339&	\underline{0.1373}/\underline{0.1452}/0.1544\\
    \bottomrule
  \end{tabular}
  \label{tab:result3}
\end{table}

\subsection*{Result 3: Robustness to  a large number of clusters}

In this experiment, we evaluate the performances of the proposed MVKMC, MVKM-ED, GKMVKM and the related methods on a data with higher ground truth number of clusters, known as 100 leaves. Regularization factor $\alpha$ is searched in $\{3, 4, 5, 6, 7, 8, 9, 10\}$ and fixed $p=2$ to GKMVKM; $\alpha=4$ to MVKM-ED, $\alpha=12$, $\beta=0.0000057$ to WMCFS; $\alpha=8$, $\beta=0.0025$ to SWVF; $\eta=10$, $\beta=5$ to TW-k-means; $\alpha=60$, $\beta=50$, $\eta=0.45$ to TW-Co-k-means. As presented in Table~\ref{tab:result4}, the proposed GKMVKM has the highest performances as compared to the other methods. Statistically, GKMVKM with $\alpha=9$ and $p=2$ show the capability to be considered as the desired parameters of finding the best pattern with promising results as compared to the other values. The results shows that the value of stabilizer $p$ by applying GKMVKM on 100 leaves data are significantly improve the clustering performances aligning on an increase $\alpha$. For this reason, we can confirm that the stabilizer $p$ of proposed GKMVKM partitions the data instance in each cluster maintained the distribution process in the good stability structure. The power densities of $p=2$ demonstrated the succeed in utilizing the effectiveness of $\alpha$ and kernel coefficients  $\beta^h$ to process the GKMVKM as a new effective way to address multi-view data.

\begin{table}
\tiny
 \caption{The Clustering Performances on 100 Leaves Dataset}
  \centering
  \begin{tabular}{lcccccc}
    \toprule
         & NMI & ARI & ACC & Recall & Precision & F-Score \\
    \midrule
   	 RMKMC	&0.2586/0.2625/0.3062&	0.0182/0.0190/0.0249	&0.0375/0.0387/0.0394&	0.7452/0.7477/0.8071&	0.0186/0.0189/0.0220&	0.0362/0.0369/0.0427\\
	WMCFS&	0/0/0	&0/0/0&	0.0100/0.0100/0.0100&	0/0/0	&0.0094/0.0094/0.0094&	0.0186/0.0186/0.0186\\
	SWVF&	0.7998/0.8209/0.8533&	0.4565/0.5053/0.5864	&0.5638/0.6158/0.7013&	0.5054/0.5656/0.6363&	0.4236/0.4652/0.5509&	0.4620/0.5104/0.5906\\
	TW-k-means&	0.8272/0.8437/0.8526&	0.4927/0.5457/0.5824&	0.5856/0.6419/0.6794&	0.5896/0.6238/0.6431&	0.4313/0.4931/0.5459	&0.4982/0.5505/0.5866\\
	TW-Co-k-means&	0.8226/0.8336/0.8435&	0.5034/0.5298/0.5660&	0.5938/0.6284/0.6719&	0.5717/0.5981/0.6293&	0.4477/0.4835/0.5294&	0.5086/0.5346/0.5704\\
	MVKMC&	0.8704/0.8859/0.9017&	0.5561/0.6342/0.6816&	0.6544/0.6994/0.7462&	0.6698/0.7190/0.7573&	0.4686/0.5741/0.6345	&0.5611/0.6380/0.6848\\
	MVKM-ED&	0.9080/0.9181/0.9272	&\textbf{0.6977}/0.7178/0.7573&	\underline{0.7256}/0.7600/0.7981	&0.7735/0.8060/0.8332	&\textbf{0.6203}/0.6521/\textbf{0.7083}	&\textbf{0.7009}/0.7208/0.7597\\
	GKMVKM 	&0.8812/0.8968/0.9128&	0.6192/0.6712/0.7208&	0.6900/0.7300/0.7744&	0.7208/0.7554/0.7891&	0.5415/0.6097/0.6683&	0.6233/0.6746/0.7237\\
	GKMVKM&	0.9007/0.9151/0.9259&	0.6805/0.7095/0.7474&	\emph{0.7194}/0.7521/0.8031&	0.7588/0.7986/0.8203&	0.6091/0.6435/0.6968&	0.6838/0.7126/0.7499\\
	GKMVKM 	&0.9115/0.9206/0.9349&	0.6866/0.7186/0.7621	&0.7163/0.7569/0.8119&	0.7871/0.8116/0.8405&	0.6129/0.6497/0.7030&	0.6899/0.7215/0.7645\\
	GKMVKM 	&0.9111/0.9225/0.9324&	0.6874/0.7224/0.7558	&0.7113/0.7596/0.8169&	0.7889/0.8157/0.8368&	\emph{0.6134}/0.6531/0.6954&	0.6907/0.7252/0.7583\\
	GKMVKM	&\emph{0.9126}/\emph{0.9237}/\emph{0.9372}&	\emph{0.6899}/\emph{0.7238}/\underline{0.7654}	&0.7188/\emph{0.7604}/\underline{0.8175}&	\underline{0.7905}/\emph{0.8182}/\emph{0.8477}&	\underline{0.6156}/\emph{0.6538}/0.7017&	\emph{0.6932}/\emph{0.7267}/\underline{0.7678}\\
	GKMVKM&	\underline{0.9131}/\textbf{0.9253/0.9391}&	\underline{0.6910}/\textbf{0.7270/0.7696}&	0.7175/\textbf{0.7630/0.8188}&	\emph{0.7893}/\textbf{0.8218}/\underline{0.8512}&	0.6012/\textbf{0.6567}/\underline{0.7062}&	\underline{0.6943}/\textbf{0.7299/0.7719}\\
	GKMVKM 	&\textbf{0.9149}/\underline{0.9248/0.9390}&	0.6842/\underline{0.7256}/\emph{0.7651}	&\textbf{0.7263}/\underline{0.7620}/\emph{0.8163}&	\textbf{0.7956}/\underline{0.8201}/\textbf{0.8535}&	0.5934/\underline{0.6556}/\emph{0.7050}&	0.6876/\underline{0.7285}/\emph{0.7675}\\
	GKMVKM &	0.6621/0.6874/0.7177&	0.2828/0.3269/00.3960	&0.4743/0.5136/0.5771&	0.3423/0.4102/0.4735&	0.2566/0.3091/0.3898&	0.3084/0.3519/0.4170\\
    \bottomrule
  \end{tabular}
  \label{tab:result4}
\end{table}

\subsection*{Result 4: Robustness to  a  higher number of dimensionalities}

For ORL-Face data, we choose and fix $\alpha=2$ and $p=3$ to GKMVKM, $\alpha=8$ to MVKM-ED, $\alpha=12$, $\beta=0.0000057$ to WMCFS; $\alpha=8$, $\beta=0.0025$ to SWVF; $\eta=10$, $\beta=5$ to TW-k-means; $\alpha=60$, $\beta=50$, $\eta=0.45$ to TW-Co-k-means. The simulation results are reported in Table~\ref{tab:result5} over 50 times in each setting algorithms. We observe that the exponential distance in proposed MVKM-ED can achieve satisfaction clustering performances compared to the proposed GKMVKM which means that the exponential distance as a rectified Euclidean norm-distance-based approach robust to high dimensionalities of one multi-view data.

\begin{table}
\tiny
 \caption{The Clustering Performances on ORL-Face  Dataset}
  \centering
  \begin{tabular}{l c c c c c c}
    \toprule
         & NMI & ARI & ACC & Recall & Precision & F-Score \\
    \midrule
    	RMKMC&	0.0000/0.1399/0.7847&	0.0000/0.0733/0.4580&	0.0250/0.1232/0.6150&	0.0000/0.5072/0.5456&	0.0226/0.0831/0.4230&	0.0441/0.1122/0.4716\\
	WMCFS&	\underline{0.7411}/0.7595/0.7811&	\underline{0.3495}/0.3998/0.4439&	\emph{0.4825}/0.5407/0.5925&	0.4667/0.5082/0.5600&	\underline{0.2966}/0.3522/0.4063&	\underline{0.3677}/0.4158/0.4581\\
	SWVF&	0.7250/0.7631/0.7928&	0.3299/0.4058/0.4807&	\underline{0.4900}/0.5429/0.6075&	0.4583/0.5199/0.5717&	0.2813/\emph{0.3557}/0.4405&	0.3486/0.4218/0.4939\\
	TW-k-means&	0.6959/0.7471/0.7885&	0.2941/0.3853/0.4774&	0.4575/0.5360/0.6100	&0.4033/0.4904/0.5728	&0.2565/0.3408/0.4353&	0.3136/0.4017/0.4907\\
	TW-Co-k-means&	\emph{0.7361}/\underline{0.7756}/0.7997&	\emph{0.3464}/\emph{0.4245}/\emph{0.4903}&	0.4750/\underline{0.5575}/\emph{0.6325}	&0.4750/0.5403/\emph{0.6028}&	\emph{0.2957}/\underline{0.3720}/\emph{0.4497}&	\emph{0.3645}/\underline{0.4399}/\emph{0.5032}\\
	MVKMC&	0.7353/0.7689/\emph{0.8076}&	0.3408/\underline{0.4153}/\underline{0.5037}&	0.4650/\emph{0.5554}/\underline{0.6500}&	\emph{0.4761/0.5328}/0.5944	&0.2806/0.3633/\underline{0.4589}	&0.3599/\emph{0.4311}/\underline{0.5163}\\
	MVKM-ED&	\textbf{0.7938/0.8226/0.8610}&	\textbf{0.4508/0.5194/0.6209}&	\textbf{0.5625/0.6322/0.7000}&	\textbf{0.6017/0.6304/0.6911}&	\textbf{0.3777/0.4611/0.5882}&	\textbf{0.4661/0.5320/0.6300}\\
	GKMVKM&	0.7304/\emph{0.7737}/\underline{0.8084}&	0.3090/0.3969/0.4775&	0.4750/0.5503/0.6250&	\underline{0.5044/0.5737/0.6239}&	0.2451/0.3256/0.4142&	0.3299/0.4144/0.4915\\
    \bottomrule
  \end{tabular}
  \label{tab:result5}
\end{table}

\subsection*{Result 5: Parameter analysis comparison with other MVC}

Another approach to measure the robustness of the rectified multi-view k-means using Gaussian-kernel is by assessing its performances in recognizing a human activity through a video. The UWA3D is the smallest dataset as compared to another dataset throughout this paper in terms of number of views. Thus, in this experiment we are decided to set the same parameters as the previous experiment to the comparison algorithms. For clarity purposes, we include changing performances of the proposed GKMVKM under different setting of parameters such as the power density parameters p is trained in $\{3, 5, 7, 9\}$ and exponent parameter $\alpha \in \{2, 3\}$. Statistically, the proposed MVKM-ED with $\alpha=6$ can be considered as the most promising clustering algorithm in explaining the multi-view pattern of UWA3D data as compared to the other algorithms. From Table~\ref{tab:result6}, we also pointed out that the proposed GKMVKM produces better clustering performance when $\alpha=2$ and $p=9$. Noted that the bold face numbers are indicated the best performance, the underline means the second-best score, and the italic clause the third-best performance.

\begin{table}
\tiny
 \caption{The Clustering Performances on UWA3D Dataset}
  \centering
  \begin{tabular}{lcccccc}
    \toprule
         & NMI & ARI & ACC & Recall & Precision & F-Score \\
    \midrule
    	RMKMC&	-/-/-&	-/-/-	&0.0533/0.0533/0.0533	&-/-/-	&0.0325/0.0325/0.0325&	0.0630/0.0630/0.0630\\
	WMCFS&	-/-/-	&-/-/-	&0.0533/0.0533/0.0533	&-/-/-	&0.0325/0.0325/0.0325	&0.0630/0.0630/0.0630\\
	SWVF&	0.7116/0.7568/\emph{0.7854}&	0.3070/0.4214/\emph{0.4909}&	0.4822/\emph{0.5594}/\underline{0.6206}&	0.4098/\underline{0.5247/0.6037}	&0.2517/0.3853/\underline{0.4656}	&0.3373/0.4432/\emph{0.5090}\\
	TW-k-means&	0.7002/0.7341/0.7789&	0.2776/0.3762/0.4739&	0.4625/0.5291/\emph{0.5929}&	0.3905/0.4782/0.5767	&0.2196/0.3456/0.4449	&0.3109/0.3998/0.4929\\
	TW-Co-k-means&	\emph{0.7217}/\textbf{0.7646}/\textbf{0.7915}&	\underline{0.3364}/\textbf{0.4376}/\textbf{0.5019	}&\underline{0.4941/0.5739}/\textbf{0.6364}&	\emph{0.4436}/\textbf{0.5347}/\textbf{0.6297}&	\underline{0.2842}/\underline{0.4027}/\textbf{0.4799}	&\underline{0.3641}/\textbf{0.4584}/\textbf{0.5209}\\
	MVKMC	&\underline{0.7248}/\emph{0.7596}/\underline{0.7859}&	\emph{0.3225}/\emph{0.4230}/\underline{0.4917}&	\emph{0.4901}/0.5584/\textbf{0.6364}&	\underline{0.4571}/\emph{0.5236}/\emph{0.5988}	&\emph{0.2670/0.3878/0.4476}&	\emph{0.3515/0.4446}/\underline{0.5107}\\
	MVKM-ED&	\textbf{0.7500}/\underline{0.7609}/0.7821&	\textbf{0.3864}/\underline{0.4331}/0.4835&	\textbf{0.5375}/\textbf{0.5818}/0.6206&	\textbf{0.4677}/0.5179/0.5844	&\textbf{0.3401}/\textbf{0.4049}/0.4475&	\textbf{0.4094}/\underline{0.4539}/0.5027\\
	GKMVKM &	0.6621/0.6845/0.7152&	0.2786/0.3249/0.3888	&0.4704/0.5112/0.5771&	0.3491/0.4040/0.4812&	0.2652/0.3092/0.3814&	0.3052/0.3498/0.4102\\
	GKMVKM &	0.6644/0.6870/0.7177&	0.2826/0.3267/0.3960	&0.4704/0.5125/0.5771&	0.3491/0.4081/0.4764&	0.2528/0.3097/0.3898&	0.3090/0.3516/0.4170\\
	GKMVKM &	0.6611/0.6864/0.7177&	0.2828/0.3265/0.3960	&0.4704/0.5137/0.5771&	0.3423/0.4084/0.4764&	0.2566/0.3092/0.3898&	0.3084/0.3514/0.4170\\
	GKMVKM &	0.6621/0.6874/0.7177&	0.2828/0.3274/0.3960	&0.4783/0.5136/0.5771&	0.3423/0.4097/0.4764&	0.2566/0.3099/0.3898&	0.3084/0.3523/0.4170\\
	GKMVKM &	0.6387/0.6655/0.6924&	0.2610/0.2921/0.3407	&0.4466/0.4887/0.5415&	0.3211/0.3671/0.4166&	0.2467/0.2815/0.3455&	0.2891/0.3182/0.3632\\
	GKMVKM &	0.6151/0.6413/0.6649&	0.2156/0.2564/0.3030	&0.4150/0.4593/0.5178&	0.2662/0.3180/0.3799&	0.2204/0.2559/0.3115&	0.2433/0.2833/0.3276\\
	GKMVKM &0.6141/0.6364/0.6568&	0.2084/0.2496/0.2918	&0.4071/0.4541/0.5020&	0.2604/0.3111/0.3549&	0.2140/0.2497/0.2868&	0.2369/0.2767/0.3172\\

    \bottomrule
  \end{tabular}
  \label{tab:result6}
\end{table}    

\section{Conclusions}

In this paper, novel rectified multi-view k-means clustering algorithms are proposed to cluster multi-view data. In MVKM-ED, an exponent Gaussian-kernel-based-distance approach is defined as a new kernel under Euclidean norm alignment to compute the distance between data points $x_i^h$ and its cluster centers $a_k^h$. Moreover, under a stabilizer $p$, the local distance is designed to MVKM-ED, named as GKMVKM. Precisely, the new exponent Gaussian-kernel-based-distance approach under Euclidean norm can be used and robust to cluster a large number of ground truth of clusters, large number of observations, and a high dimensional of multiple resources data. In the context of extensive experiments gather on synthetic and five real-world datasets, our proposed MVKM-ED and GKMVKM representing the superiority compared to the other state-of-the-art multi-view k-means clustering algorithms. Notice that the $p$ stabilizer parameter on GKMVKM play a critical role in applying exponential Gaussian-kernel distance as it can resulting clustering performances in a huge changes.

\bibliographystyle{unsrt}  


\end{document}